%% file: main_agent.tex
\documentclass[10pt,twocolumn,letterpaper]{article}

\usepackage{cvpr}              
\usepackage[accsupp]{axessibility}

\input{preamble}

\definecolor{cvprblue}{rgb}{0.21,0.49,0.74}
\usepackage[pagebackref,breaklinks,colorlinks,citecolor=cvprblue]{hyperref}

\def\paperID{13846} 
\def\confName{CVPR}
\def\confYear{2024}

\title{ChatScene: Knowledge-Enabled Safety-Critical Scenario Generation for Autonomous Vehicles}
\author{
Jiawei Zhang\\
UIUC\\
\blackhref{mailto:jiaweiz7@illinois.edu}{jiaweiz7@illinois.edu}
\and
Chejian Xu\\
UIUC\\
\blackhref{mailto:chejian2@illinois.edu}{chejian2@illinois.edu}
\and
Bo Li\\
UChicago \& UIUC\\
\blackhref{mailto:bol@uchicago.edu}{bol@uchicago.edu}
}
\begin{document}
\maketitle
\begin{abstract}
\vspace{-10pt}

We present \name, a Large Language Model (LLM)-based agent that leverages the capabilities of LLMs to generate safety-critical scenarios for autonomous vehicles. Given unstructured language instructions, the agent first generates textually described traffic scenarios using LLMs. These scenario descriptions are subsequently broken down into several sub-descriptions for specified details such as behaviors and locations of vehicles.
The agent then distinctively transforms the textually described sub-scenarios into domain-specific languages, which then generate actual code for prediction and control in simulators, facilitating the creation of diverse and complex scenarios within the CARLA simulation environment. A key part of our agent is a comprehensive knowledge retrieval component, which efficiently translates specific textual descriptions into corresponding domain-specific code snippets by training a knowledge database containing the scenario description and code pairs. Extensive experimental results underscore the efficacy of \name in improving the safety of autonomous vehicles. 
For instance, the scenarios generated by \name show a $15\%$ increase in collision rates compared to state-of-the-art baselines when tested against different reinforcement learning-based ego vehicles. Furthermore, we show that by using our generated safety-critical scenarios to fine-tune different RL-based autonomous driving models, they can achieve a $9\%$ reduction in collision rates, surpassing current SOTA methods. \name effectively bridges the gap between textual descriptions of traffic scenarios and practical CARLA simulations, providing a unified way to conveniently generate safety-critical scenarios for safety testing and improvement for AVs. The code is available at~\url{https://github.com/javyduck/ChatScene}.
\end{abstract}

\vspace{-7mm}
\section{Introduction}

\begin{figure*}[ht]
    \centering
    \includegraphics[width=\textwidth]{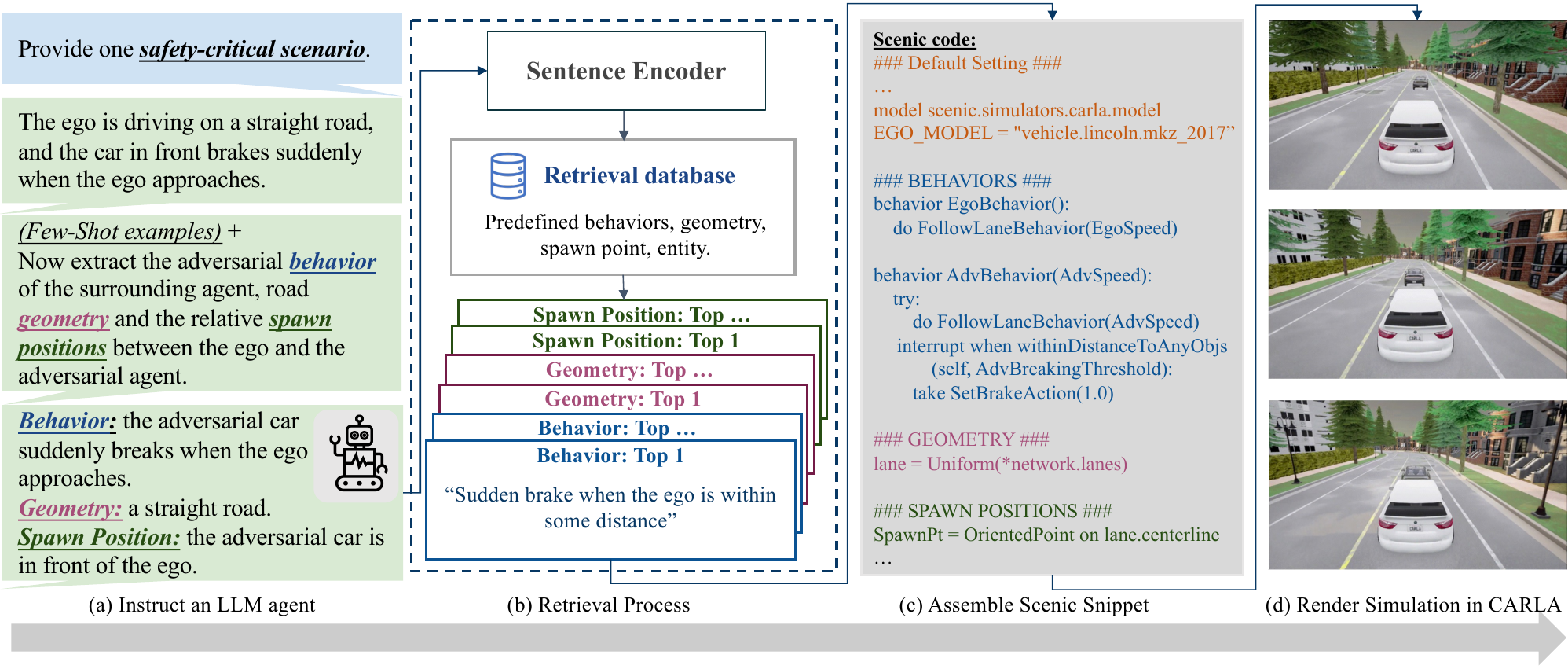}
    \vspace{-8mm}
    \caption{\small Overview of our LLM-based knowledge-enabled safety-critical scenario generation agent~\name.}
    \label{fig:pipeline}
    \vspace{-7mm}
\end{figure*}

Although machine learning (ML), particularly deep neural networks (DNNs), has shown remarkable performance across a myriad of applications such as image recognition~\citep{he2016deep}, natural language processing~\citep{devlin2018bert}, and healthcare~\citep{erickson2017machine}, they also exhibit a surprising susceptibility to subtle and adversarial perturbations. These perturbations can yield erroneous predictions~\citep{biggio2013evasion,szegedy2014intriguing}, posing potentially fatal consequences in safety-critical applications like Autonomous Driving (AD)~\citep{kong2020physgan,cao2019adversarial}. For example, by attaching seemingly innocuous stickers to a \emph{Stop Sign} in the real world, an autonomous vehicle (AV) can readily misinterpret it as a \emph{Speed Limit 80 Sign}~\citep{eykholt2018robust}, which can lead to some hazardous driving behaviors and potential accidents. 

Therefore, given the potential for such adversarial manipulations, it is crucial that AVs undergo exhaustive testing across all conceivable safety-critical scenarios to ensure their safe and reliable operation before large-scale deployment. However, traditional real-world testing is not only prohibitively expensive but also demands extensive data collection, often requiring vehicles to be driven hundreds of millions of miles to accumulate sufficient safety-critical scenarios. Consequently, the generation of simulated scenarios for testing has been increasingly adopted as a cost-effective and efficient alternative. 

For instance, Wachi \etal~\cite{wachi2019failure} employ multi-agent reinforcement learning to train adversarial vehicles, aiming to expose the vulnerabilities in rule-based driving algorithms within the CARLA platform~\cite{dosovitskiy2017carla}. Chen \etal~\cite{chen2021adversarial} instead focus on generating adversarial scenarios for lane-changing maneuvers using ensemble deep reinforcement learning techniques, and Feng \etal~\cite{feng2021intelligent} further offer a highway-driving simulation that incorporates further scenarios such as \emph{Cut Following} and \emph{Cut-in}. Nevertheless, a key challenge remains: these methods are limited to only a narrow range of safety-critical scenarios, which may still fall short of encompassing the complexity of real-world situations. 

Meanwhile, the emergence of LLMs, trained with vast amounts of data from the internet and encompassing billions of parameters, has demonstrated a remarkable aptitude for capturing human knowledge~\cite{brown2020language, chowdhery2022palm, touvron2023llama, zheng2023judging}. These models have established themselves as effective tools for the extraction of knowledge. For instance, LLMs begin to play a pivotal role in pedagogical processes~\cite{kasneci2023chatgpt}, and they are becoming instrumental in synthesizing clinical knowledge to support medical practice~\cite{singhal2022large}. The legal field also benefits from LLMs, with tools like Chatlaw~\cite{cui2023chatlaw} interpreting legal regulations and judicial decisions, while in the financial sector, models such as BloombergGPT~\cite{wu2023bloomberggpt} are being harnessed to decode complex economic data.

This naturally leads to several compelling questions: \emph{Is it possible to build an LLM-based agent that automatically generates safety-critical scenarios, capturing a broader and more intricate array of descriptions? Moreover, can the agent automatically convert these textual descriptions into real simulations to bolster the diversity and comprehensiveness of scenarios available for AV testing?}

Addressing the initial query of generating safety-critical driving scenario descriptions vian LLM agent is a process that is relatively straightforward; one can simply prompt the model with requests such as,~\emph{``Provide some descriptions for safety-critical driving scenarios.''} While for the second question, the recent advancements of~\emph{Scenic}~\cite{fremont2019scenic, fremont2022scenic}, a domain-specific probabilistic programming language, allow for the scripting of scenes within CARLA~\cite{dosovitskiy2017carla} using syntax akin to Python, opens up two promising research directions: first, the possibility of guiding LLMs to autonomously script in Scenic, and second, the potential for finetuning a language-to-code model, such as~\emph{CodeGen}~\cite{nijkamp2022codegen}, to craft Scenic code derived from textual scenario descriptions. Nonetheless, these methods often encounter obstacles, such as the generation of non-executable code or calls to APIs that do not exist within Scenic, primarily due to the scarcity of available code examples for training.

To address the challenges of direct code generation by large language models (LLMs), we instead adopt an indirect approach that leverages LLMs to first curate a retrieval database comprising Scenic code snippets. These snippets encapsulate fundamental elements of driving scenarios, such as adversarial behaviors of surrounding vehicles, road geometries, etc. The details of the construction process will be introduced in ~\Cref{sec:database}. Then, during the evaluation phase, as illustrated in \Cref{fig:pipeline}, our agent~\name maps the description into the corresponding simulation through a four-step process:

\begin{enumerate}[leftmargin=2em, label=\roman*.]
    \item Given instruction from the user, \name generates a natural language description of a safety-critical scenario leveraging the intrinsic wide knowledge in LLMs.
    \item \name further parses this description, extracting detailed characteristics that align with critical scenario components, such as the adversarial behaviors of surrounding vehicles.
    \item \name then encodes these characteristics into embeddings to retrieve the corresponding Scenic code snippets from our pre-constructed database.
    \item Finally, the retrieved snippets are assembled into a complete and executable Scenic script, which is capable of enacting the described scenario within the CARLA simulation environment.
\end{enumerate}

\begin{figure*}[ht]
    \centering
    \includegraphics[width=\textwidth]{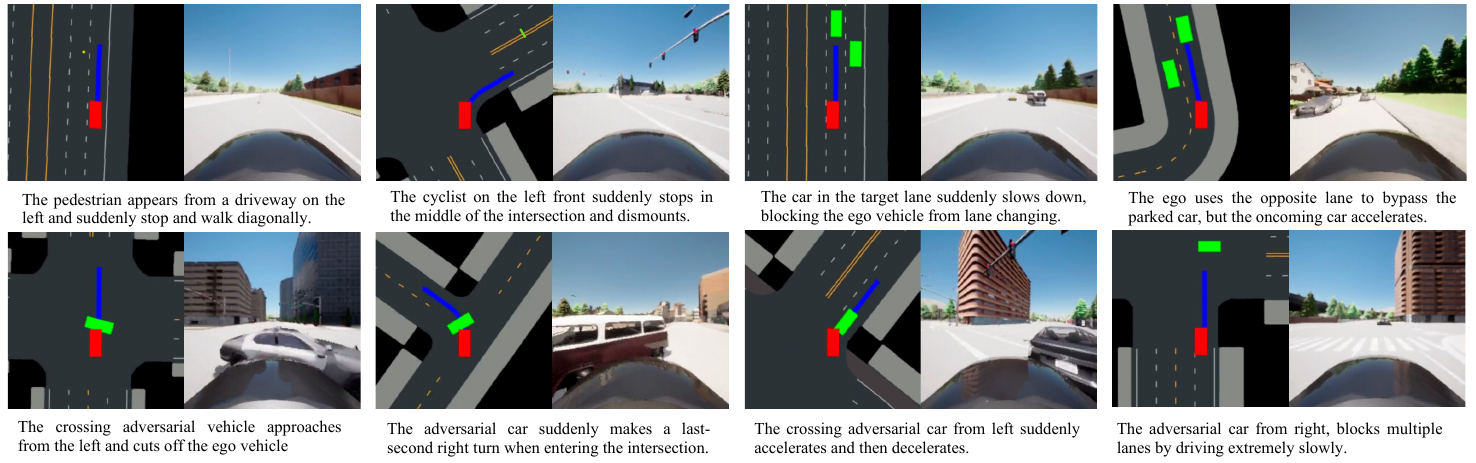}
    \vspace{-7mm}
    \caption{\small Eight selected text-simulation cases using our agent~\name, the descriptions here are shortened for clarity.}
    \label{fig:simulation}
    \vspace{-7mm}
\end{figure*}

\vspace{-1mm}
For a further quantitative analysis, we utilize \name to produce a range of text descriptions for safety-critical scenarios. These narratives are then processed by our framework to generate simulations, which are then evaluated using the Safebench platform~\cite{xu2022safebench}. Within this platform, the ego vehicle is controlled by a model trained under reinforcement learning, while our scenario generation agent aims to manage the adversarial objects (e.g., pedestrian, cyclist, or vehicle) surrounding it. Some examples of text-to-simulation mappings are provided in~\Cref{fig:simulation}, illustrating the practical application of our framework.

Our contributions can be concluded as:

\begin{itemize}[leftmargin=2em]
    \item We introduce \name, a novel LLM-based agent capable of generating safety-critical scenarios by first providing textual descriptions and then carefully transforming them into executable simulations in CARLA via Scenic programming language.
    
    \item An expansive retrieval database of Scenic code snippets has been developed. It catalogs diverse adversarial behaviors and traffic configurations, utilizing the rich knowledge stored in LLMs, which significantly augments the variety and critical nature of the driving scenarios generated. 
    
    \item In Safebench's evaluation of eight CARLA Challenge traffic scenarios~\cite{carla_challenge_2019}, our method's adversarial scenes increased the collision rate by $15\%$ compared to four state-of-the-art (SOTA) baselines, demonstrating the superior safety-critical capabilities of our framework.

    \item Subsequent experiments involving the finetuning of the ego vehicle with a subset of our generated adversarial scenarios, followed by comparative evaluations against both the remaining scenarios we created and those from established baselines, demonstrated an additional reduction in average collision rates by at least $9\%$.

    \item Our framework, in conjunction with the retrieval database, not only facilitates direct code generation but also holds potential for future adaptations in multimodal conversions, including text, image, and video, specifically for autonomous driving applications.
\end{itemize}

\vspace{-5mm}
\section{Related Work}
\vspace{-2mm}

\paragraph{Generation of safety-critical scenarios.}

The generation of safety-critical scenarios for autonomous vehicles (AVs) generally falls into three main categories. The first is~\emph{data-driven generation}~\citep{van2015automated, yang2020surfelgan, ding2020cmts, scanlon2021waymo, tan2023language}, which relies on real-world data to guide vehicle behavior. While realistic, this approach usually suffers from the scarcity and high cost of gathering pertinent data and the infrequency of genuinely risky scenarios within the collected dataset since the safety-critical scenarios usually lie on the long-tail distribution of the real-world scenario distribution. The second category,~\emph{adversarial generation}~\citep{lee2015adaptive, abeysirigoonawardena2019generating, prakash2019structured, rempe2022generating, zhang2023cat}, intentionally conducts malicious attacks against the AVs by manipulating the behaviors of surrounding vehicles, such as pedestrians or other vehicles. While effective at creating challenging environments, this method may be computationally inefficient and may lack diversity in the generated scenarios. The third approach,~\emph{knowledge-based generation}~\citep{rocklage2017automated, bagschik2018ontology, cai2020summit, klischat2020scenario}, uses predefined traffic rules or physical constraints to create scenarios. Although it is more systematic and can provide more diverse scenarios, this approach can be challenging to implement, as encoding these rules into simulations can be complex. Besides, the manually created rules are hard to cover all safety-critical situations, resulting in less risky scenarios since 
they typically do not incorporate adversarial attacks or unexpected behaviors that are crucial for testing the robustness of AVs. 

Our work, instead, synergies the advantages of the latter two categories by integrating diverse real-world knowledge rules sourced from LLMs while using Scenic to adversarially optimize parameters of the surrounding environment, such as the speed of nearby pedestrians and vehicles, to enhance the risk and complexity of the generated scenarios. 

\vspace{-5mm}
\paragraph{LLMs for autonomous driving.} LLMs have increasingly been explored for their potential in autonomous driving, particularly in their ability to interpret complex scenarios in a way that resembles human understanding. For instance, Fu~\etal~\cite{fu2023drive} utilize GPT-3.5 in a reasoning and action prompt style~\cite{yao2022react} to generate API-wrapped text descriptions of decisions made in response to observations in a highway environment. On the other hand, Xu~\etal~\cite{xu2023drivegpt4} propose to leverage a multimodal language model to interpret driving scenarios and provide corresponding descriptions with the prediction for the next control signals based on the driving video frames and human questions. Additionally, Zhong~\etal~\cite{zhong2023language} introduces the use of the LLM to transform a user's query about safety-critical scenarios into the corresponding differentiable loss function of a diffusion model to generate the query-compliant trajectories. 

Our work differs from the first two studies in that we focus primarily on the generation of safety-critical scenarios rather than using LLMs to provide descriptions or actions for specific driving situations. Besides, in contrast to the last work, our work mainly aims to create more realistic safety-critical scenario generations on a platform like CARLA and use the corresponding driving record to train or test the ego vehicle controlled trained with reinforcement learning. 

\begin{figure*}[t]
    \centering
    \includegraphics[width=\textwidth]{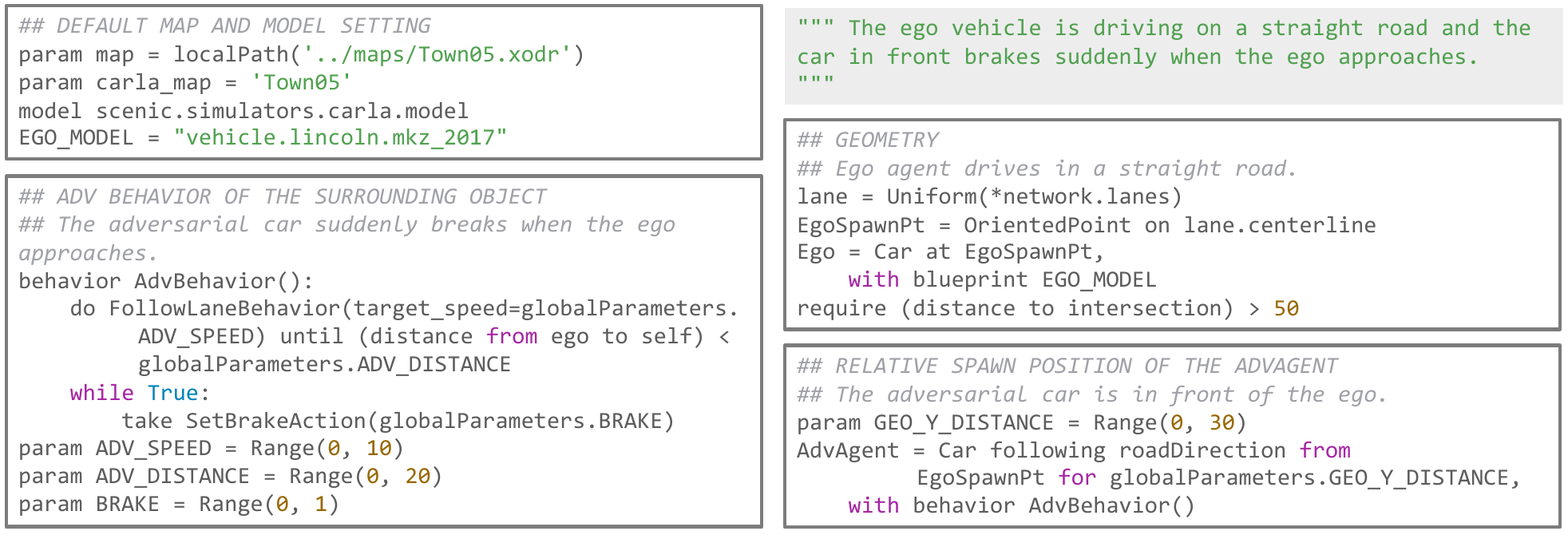}
    \vspace{-8mm}
    \caption{\small An example for the snippets in Scenic for a given safety-critical scenario description. }
    \label{fig:retrieval_code}
    \vspace{-7mm}
\end{figure*}

\vspace{-4mm}
\section{Methodology}
\label{sec:methodology}
\vspace{-2mm}

In this section, we delineate the approach of~\name, an LLM agent, for the generation of safety-critical scenarios through the application of Scenic programming language and a targeted retrieval mechanism.  We begin with a concise motivation, including definitions of key terminologies. Then, we focus on the construction of the retrieval database, and following this, we explain the process of converting text descriptions of safety-critical scenarios, sourced from either humans or LLMs, into Scenic code. Specifically, we will begin by breaking down the original text description into sub-descriptions for each component (e.g., adversarial behavior). Next, we encode these components into vectors, which will serve as keys to retrieve the corresponding snippets from our retrieval database. These snippets will then be assembled to form a comprehensive Scenic script, which is then executed to run simulations within the CARLA environment.

\vspace{-3mm}
\subsection{Motivation and Notations}
\label{sec:motivation}
\vspace{-2mm}

We notice that the Scenic~\cite{fremont2019scenic, fremont2022scenic} programming language is highly effective and flexible for rendering simulations in CARLA. A pertinent question naturally arises:~\emph{``Can we directly prompt large language models like ChatGPT to write corresponding Scenic code based on descriptions of safety-critical scenarios?''} Although this approach seems promising, it often leads to issues like generating non-compilable code or using APIs not present in the codebase. This might be due to the scarcity of Scenic examples for training the LLMs and the complexity and breadth of code generation, which can induce hallucinations in the model. 

However, upon examining Scenic code, we find that it can generally be segmented into four components as shown in~\Cref{fig:retrieval_code}: (1) default map and model settings, typically fixed; (2) definition of \textit{adversarial behavior} for surrounding vehicles; (3) the \textit{road geometry}, which also influences the ego vehicle's spawn point; and (4) the \textit{relative spawn position} of surrounding vehicles to the ego vehicle. Notably, the ego vehicle is controlled by models trained via reinforcement learning, so we don't need to define its behavior here. 

Based on this, we propose a more efficient method: collecting a database of code snippets for the last three components with corresponding descriptions instead. This approach would likely reduce hallucination and allow for flexible assembly of these snippets into a complete Scenic code. Then, during evaluation, for input descriptions like~\emph{``The ego vehicle is driving on a straight road, and the car in front brakes suddenly as the ego approaches''}, our LLM agent can first  decompose it into the sub-descriptions for each component via demonstrations easily, such as~\emph{``Behavior: the adversarial car suddenly brakes as the ego approaches''} for the behavior component. Then, our agent will do the corresponding retrieval and assemble the relevant code snippets based on the embedding of these descriptions to generate comprehensive Scenic code for simulation in CARLA. To provide a clearer understanding of our methodology, we define the following foundational concepts: 

\vspace{-5mm}
\paragraph{Route.}
A \textit{`route'} is essentially a sequence of waypoints, each marking a specific location that the vehicle is intended to pass through during its trajectory. In the CARLA simulation environment, this route represents a pre-determined path for the ego vehicle that includes both the starting and the terminal points.

\vspace{-5mm}
\paragraph{Base Scenario.}
A \textit{`base scenario'} is utilized to conceptualize a high-level adversarial driving situation. It provides an abstract framework, such as~\emph{``a straight obstacle ahead of the ego vehicle''}, without delving into specifics regarding the identity or adversarial behavior of the obstacle. This abstraction allows for a generalized approach to categorizing various driving challenges.

\vspace{-5mm}
\paragraph{Scenario.}
In contrast, a \textit{`scenario'} builds upon the \textit{`base scenario'} by infusing it with detailed attributes concerning the obstacle and its specific behaviors. For instance, the previous example~\emph{``The ego vehicle is driving on a straight road, and the car in front brakes suddenly as the ego approaches''} is a scenario derived from the foundational base scenario as shown above, which enriches the initial description by introducing specific dynamics of the adversarial situation.

\vspace{-5mm}
\paragraph{Scene.}
A \textit{`scene'} represents the practical instantiation of a \textit{`scenario'}, detailing the specifics of an adversarial event. This includes the route of the ego vehicle, the characteristics of the surrounding adversarial vehicle (\textit{e.g.}, vehicle type), and environmental context (\textit{e.g.}, buildings or traffic signals). Furthermore, it specifies parameters such as the positions, velocities, and initial placements of the agents involved. Notably, Scenic actually acts as a \textit{probabilistic} programming language as shown in~\Cref{fig:retrieval_code}; it will sample the parameters like \texttt{ADV\_SPEED} before running the corresponding simulation in CARLA. This capability enables a single Scenic script to produce various distinct scenes for the same scenario by varying parameter values like speed within a predefined range.

\vspace{-2mm}
\subsection{Construction of the Retrieval Database}
\label{sec:database}
\vspace{-2mm}

This section details our methodology for building the retrieval database, emphasizing the systematic collection and integration of Scenic code snippets.

\vspace{-5mm}
\paragraph{Collection of Snippets.}
Our snippet collection process initiates with sourcing initial examples from the Scenic repository\footnote{\url{https://github.com/BerkeleyLearnVerify/Scenic/tree/main/examples/carla}}. These examples, covering a range of adversarial behaviors and geometric configurations (e.g., straight roads, intersections), are then decomposed manually into description-snippet pairs. Utilizing this initial combined dataset, we engage LLMs for few-shot learning to generate more diverse snippets. This iterative process involves generating snippets for different components, including adversarial behaviors, geometric layouts, and relative spawn points. Each newly generated snippet is rigorously evaluated and corrected manually for its compatibility and compilability within Scenic’s varying API contexts (such as differences across agent types like pedestrians, motorcycles, and cars). In this work, we consistently use \texttt{GPT-4}\footnote{\url{https://chat.openai.com/?model=gpt-4}} to collect our snippets. Some examples of the prompts employed to generate these new snippets are provided in~\Cref{adx:prompt_snippet}.

\vspace{-5.4mm}
\paragraph{Database Construction and Query Optimization.} For every description-snippet pair, we continue to prompt~\texttt{GPT-4} to generate several more distinct rephrasings of each description, maintaining the original snippet in each pair. This approach is designed to enhance the accuracy of retrieval. The descriptions are then encoded using \texttt{Sentence-T5}\footnote{\url{https://huggingface.co/sentence-transformers/sentence-t5-large}}~\cite{ni2022sentence}, and the database construction and querying processes are facilitated by \texttt{faiss}~\cite{johnson2019billion}. We construct and manage the databases for different components,~\emph{i.e.,} adversarial behavior, road geometry, and relative spawn positions, independently.

\begin{table*}[t]
\centering
\small
    \caption{\textbf{Statistics of scenario generation on selected scenarios}. We report \textit{\collisionrate} (\collisionrateabbr), the \textit{\overallscore} (\overallscoreabbr), and the~\emph{average displacement error} (ADE)  to measure the effectiveness of different scenario generation algorithms; we test three differently trained ego vehicles, and the record herein represent the mean performance across these agents with all the scenes for the same base scenario. The last column shows the average over all the base scenarios, with bold numbers indicating the best performance among the $5$ generation algorithms. \Learningtocollideabbr: \Learningtocollide, \AdvSimabbr: \AdvSim, \CarlaGeneratorabbr: \CarlaGenerator, \AdvTrajabbr: \AdvTraj, $\uparrow$/$\downarrow$: higher/lower the better.}
    \vspace{-3mm}
    \label{tab:test_generation}
{
\resizebox{0.8\textwidth}{!}{
\setlength{\tabcolsep}{3.75pt}
    \begin{tabular}{c|c|cccccccc|c}
    \toprule
        \multirow{3}{*}{\textbf{Metric}} & \multirow{3}{*}{\textbf{Algo.}} & \multicolumn{8}{c|}{\textbf{Base Traffic Scenarios}} & \multirow{3}{*}{\textbf{Avg.}} \\
        & & \scriptsize{\makecell{Straight \\ Obstacle}} & \scriptsize{\makecell{Turning \\ Obstacle}} & \scriptsize{\makecell{Lane \\ Changing}}  & \scriptsize{\makecell{Vehicle \\ Passing}} & \scriptsize{\makecell{Red-light \\ Running}} & \scriptsize{\makecell{Unprotected \\ Left-turn}} & \scriptsize{\makecell{Right-\\ turn}} & \scriptsize{\makecell{Crossing \\ Negotiation}} & \\
        \midrule
          \multirow{5}{*}{\collisionrateabbr $\uparrow$} & \Learningtocollideabbr&0.30 & 0.09 & 0.87 & 0.83 & 0.71 & 0.69 & 0.59 & 0.58 & 0.584\\
           & \AdvSimabbr&  0.51 & 0.33 & 0.86 & 0.87 & 0.57 & 0.70 & 0.29 & 0.57 & 0.586 \\
           & \CarlaGeneratorabbr &  0.45 & 0.61 & 0.89 & 0.87 & 0.63 & 0.69 & 0.68 & 0.60 & 0.676 \\
           & \AdvTrajabbr & 0.50 & 0.31 & 0.78 & 0.82 & 0.71 & 0.68 & 0.59 & 0.62 & 0.627  \\
           & \name &  \textbf{0.89} & \textbf{0.70} & \textbf{0.95} & \textbf{0.93} & \textbf{0.79} & \textbf{0.75} & \textbf{0.78} & \textbf{0.86} & \textbf{0.831}  \\
        \midrule
          \multirow{5}{*}{\overallscoreabbr $\downarrow$} & \Learningtocollideabbr & 0.761 & 0.830 & 0.505 & 0.507 & 0.601 & 0.615 & 0.548 & 0.588 & 0.619 \\
           & \AdvSimabbr &  0.673 & 0.707 & 0.507 & 0.490 & 0.675 & 0.607 & 0.705 & 0.593 & 0.620 \\
           & \CarlaGeneratorabbr & 0.698 & 0.567 & 0.489 & 0.490 & 0.641 & 0.613 & 0.505 & 0.579 & 0.573 \\
           & \AdvTrajabbr &0.668 & 0.714 & 0.538 & 0.505 & 0.607 & 0.620 & 0.545 & 0.569 & 0.596 \\
            & \name & \textbf{0.470} & \textbf{0.522} & \textbf{0.434} & \textbf{0.440} & \textbf{0.537} & \textbf{0.560} & \textbf{0.474} & \textbf{0.421} & \textbf{0.482}  \\
  \midrule
          \multirow{5}{*}{ADE $\uparrow$} & \Learningtocollideabbr & 0.467 & 0.178 & 0.330 & 0.000 & 0.866 & 0.585 & 1.476 & 0.805 & 0.588\\
           & \AdvSimabbr& 0.291 & 0.073 & 0.242 & 0.000 & 0.365 & 0.754 & 0.628 & 0.398 & 0.344 \\
           & \CarlaGeneratorabbr & 0.348 & 1.668 & 0.410 & 0.282 & 0.324 & 0.338 & 0.385 & 0.299 & 0.507\\
           & \AdvTrajabbr & 0.683 & 1.236 & 3.762 & 0.000 & 1.931 & 1.720 & 1.921 & 2.301 & 1.694 \\
           & \name &  \textbf{4.398} & \textbf{4.063} & \textbf{5.706} & \textbf{7.383} & \textbf{3.848} & \textbf{3.740} & \textbf{3.613} & \textbf{3.784} & \textbf{4.567}\\
        \bottomrule
    \end{tabular}
  }
}
\vspace{-6mm}
\end{table*}

\vspace{-2mm}
\subsection{Safety-Critical Scenario Generation}
\label{sec:pipeline}
\vspace{-2mm}

Upon the completion of the retrieval database, our LLM agent, \name can now first generate a variety of descriptions for safety-critical scenarios and then convert them into the corresponding simulation via Scenic code during the evaluation. The detailed process is shown as follows:

\vspace{-1mm}

\begin{enumerate}[leftmargin=2em, label=\roman*.]
\item \textbf{Instruct the LLM agent}: We start by instructing our LLM agent to generate potential adversarial scene descriptions. An example of such a query is: \emph{``Provide a description of a safety-critical scenario where the ego vehicle is driving on a straight road.''}

\item \textbf{Description Extraction for Each Component}: To guarantee structured output for each scenario component, our LLM agent will then automatically employ a set of few-shot examples to guide it in generating organized sub-descriptions, which are formatted as ``\textit{Behavior: ...$\backslash$n Geometry: ...$\backslash$n Spawn Position: ...}'' as illustrated in~\Cref{fig:pipeline} (a). Consequently, the agent can then efficiently extract the corresponding descriptions for each component using regular expressions. The prompts for the extraction are detailed in~\Cref{adx:prompt_description}.

\item \textbf{Retrieving Scenic Code Snippets}: After extracting descriptions, our agent will utilize the \texttt{Sentence-T5} model to encode them. The embeddings will serve as keys for retrieving the relevant Scenic code snippets for each component, as shown in \Cref{fig:pipeline} (b) and (c).

\item \textbf{Scenario Rendering and Evaluation}: The Scenic code snippets are then assembled into a complete script and executed to run simulations in CARLA as demonstrated in~\Cref{fig:pipeline} (d). More text and simulation pairs are shown in~\Cref{fig:simulation}. Then, different parameter values, such as \texttt{ADV\_SPEED} set between \texttt{[0, 10]} and \texttt{ADV\_DISTANCE} between \texttt{[0, 20]}, will be sampled for collecting multiple scenes. The comprehensive details such as position, speed, acceleration, and collision information for all vechiles in each frame.

\item \textbf{Refinement of Collision-Prone Parameters}: To enhance the generation of the scenes that lead to the collision of the ego vehicle, the sampling ranges will be dynamically adjusted based on the information gained from previously collected data. Specifically, we will keep recording the parameters associated with collision cases and simply assume that parameters leading to collisions roughly align with a Gaussian distribution \( \mathcal{N}(\mu, \sigma^2) \). Consequently, the sampling range will then be adjusted to \([\mu - \sigma, \mu + \sigma]\) for subsequent simulations. This iterative strategy has proven effective in increasing the probability of generating more collision-prone scenes. In the end, the most adversarially significant scenes will be kept for testing on each scenario.
\end{enumerate}

\vspace{-1mm}
The complete set of simulations along with comprehensive statistics are released for all involved vehicles. This data will support future research in the bidirectional conversions among text, image, and video for autonomous driving.

\vspace{-4mm}
\section{Experiment}
\label{sec:exp}
\vspace{-2mm}

In this section, we conduct a quantitative evaluation of our agent in generating safety-critical scenarios. Our assessment is twofold: First, we test the actual safety-critical nature of the scenes produced by our agent, specifically their potential to provoke collisions involving the ego vehicle. Second, we evaluate the performance of the ego vehicle, which has undergone adversarial retraining using scenarios generated by our agent, to ascertain whether these scenarios contribute significantly to enhancing the robustness of ego vehicle.

\vspace{-2mm}
\subsection{Setup}
\label{sec:setup}
\vspace{-2mm}

In this work, to simulate autonomous driving, we control the ego vehicle with a reinforcement learning-based model and employ Scenic to guide the surrounding adversarial vehicle. Besides, for a more flexible and convenient implementation, we integrate Scenic into the Safebench platform~\cite {xu2022safebench} and conduct all the evaluations on Safebench.

\vspace{-6mm}
\paragraph{AD algorithms.} Safebench provides three prominent deep RL methodologies to train our ego vehicle. These are: Proximal Policy Optimization (PPO)~\citep{schulman2017proximal}, an on-policy stochastic algorithm; Soft Actor-Critic (SAC)~\citep{haarnoja18b}, an off-policy stochastic technique; and Twin Delayed Deep Deterministic Policy Gradient (TD3)~\citep{fujimoto18a}, a deterministic off-policy approach. The observation of the ego encompasses four essential dimensions: the distance to the next waypoint, longitudinal speed, angular speed, and a detection signal for front-facing vehicles.

\vspace{-5mm}
\paragraph{Baselines.} We employ two main categories of scenario-generation techniques for evaluation: Adversary-based and Knowledge-based. Adversary-based approaches, like \textbf{\Learningtocollide (\Learningtocollideabbr)}~\cite{ding2020learning} and  \textbf{\AdvSim (\AdvSimabbr)}~\cite{wang2021advsim}, challenge AD systems by altering initial poses of agents or perturbing trajectories. Knowledge-based approaches, such as \textbf{\CarlaGenerator (\CarlaGeneratorabbr)}~\cite{scenariorunner} and \textbf{\AdvTraj (\AdvTrajabbr)}~\cite{zhang2022adversarial}, focus on scenarios adhering to real-world traffic rules and physical principles.


\begin{table*}[t]
\small
    \centering
    \caption{\small \textbf{Diagnostic Report:} This report presents the average test results conducted using three distinct ego vehicles across eight base scenarios. These tests are evaluated on three different performance levels for each scenario generation algorithm, offering a comprehensive overview of agent efficacy. \collisionrateabbr: \collisionrate, \runredlightabbr: \runredlight, \runstopsignabbr: \runstopsign, \outofroadabbr: \outofroad, \followrouteabbr: \followroute, \routecompletionabbr: \routecompletion, \timespentabbr: \timespent, \accelerationabbr: \acceleration, \yawvelocityabbr: \yawvelocity, \laneinvasionabbr: \laneinvasion, \overallscoreabbr: \overallscore, $\uparrow$/$\downarrow$: higher/lower the better.}
    \vspace{-3mm}
    \label{tab:diagnostic}
{
    \resizebox{0.8\textwidth}{!}{
    \setlength{\tabcolsep}{3.75pt}
    \begin{tabular}{c|cccc|ccc|ccc|c}
    \toprule
    \multirow{2}{*}{\textbf{Algo.}} & \multicolumn{4}{c|}{\textbf{Safety Level}} & \multicolumn{3}{c|}{\textbf{Functionality Level}} & \multicolumn{3}{c|}{\textbf{Etiquette Level}} & \multirow{2}{*}{\textbf{\overallscoreabbr $\downarrow$}} \\
    & \collisionrateabbr $\uparrow$ & \runredlightabbr $\uparrow$ & \runstopsignabbr $\uparrow$ & \outofroadabbr $\uparrow$ & \followrouteabbr $\downarrow$ & \routecompletionabbr $\downarrow$ & \timespentabbr $\uparrow$ & \accelerationabbr $\uparrow$ & \yawvelocityabbr $\uparrow$ & \laneinvasionabbr $\uparrow$ & \\
    \midrule
      \Learningtocollideabbr& 0.584 & \textbf{0.326} & 0.158 & 0.032 & 0.894 & 0.731 & 0.216 & 0.211 & 0.243 & 0.112 & 0.619\\
        \AdvSimabbr& 0.586 & 0.300 & 0.160 & 0.025 & 0.891 & 0.745 & 0.261 & 0.203 & 0.245 & 0.127 & 0.620  \\
        \CarlaGeneratorabbr & 0.676 & 0.313 & \textbf{0.161} & \textbf{0.036} & 0.890 & 0.741 & 0.244 & 0.215 & 0.243 & 0.131 & 0.573 \\
     \AdvTrajabbr & 0.627 & 0.312 & 0.158 & 0.028 & 0.893 & 0.726 & \textbf{0.279} & 0.219 & 0.248 & 0.137 & 0.596 \\
     \name & \textbf{0.831} & 0.179 & 0.143 & 0.035 & \textbf{0.833} & \textbf{0.544} & 0.223 & \textbf{0.705} & \textbf{0.532} & \textbf{0.243} & \textbf{0.482}  \\
    \bottomrule
    \end{tabular}
    }
}
\vspace{-6mm}
\end{table*}

\vspace{-5mm}
\paragraph{Metrics.} Evaluation under Safebench encompasses three categories: \textit{Safety level} (including collision rate and adherence to traffic signals), \textit{Functionality level} (route adherence and completion), and \textit{Etiquette level} (smoothness of driving and lane discipline). Our focus primarily lies on the \emph{collision rate} and a composite \emph{\overallscore (\overallscoreabbr)}, with the latter aggregating all metrics, further details are deferred to~\Cref{apx:metric}. We also leverage the \emph{average displacement error} (ADE) to measure scene diversity generated by each algorithm. Specifically, it is calculated as the average of the mean of the Euclidean distances between the positions of the adversarial objects at each corresponding time step across the trajectories for each pair of scenes generated for the same scenario.

\vspace{-2mm}
\subsection{Safety-Critical Scenarios Generation}
\label{exp:generation}
\vspace{-2mm}
This section explores the capabilities of our agent in generating the most adversarial safety-critical scenarios when compared to the baselines.

\vspace{-3mm}
\paragraph{Experiment Setup.}
Following Safebench~\citep{xu2022safebench}, we leverage a surrogate ego vehicle trained using SAC to choose the most challenging scenes generated by various methods. Following this, we then evaluate these selected adversarial scenes using three distinct ego vehicles trained via SAC, PPO, and TD3. This assesses the effectiveness and generality of different algorithms for generating safety-critical scenarios.

\vspace{-3mm}
\paragraph{Base Scenario and Route.} We adopt eight key base traffic scenarios from the Carla Challenge~\citep{carla_challenge_2019}, whose texts are summarized from the NHTSA report~\citep{najm2007pre}, each with $10$ diverse routes for the ego vehicle. These base scenarios are: \emph{Straight Obstacle}, \emph{Turning Obstacle}, \emph{Lane Changing Vehicle}, \emph{Passing Red-light Running}, \emph{Unprotected Left-turn}, \emph{Right-turn}, and \emph{Crossing Negotiation}. 

\vspace{-3mm}
\paragraph{Scenario.} In contrast to the baseline methods, which offer only one single scenario per base scenario, our approach demonstrates greater diversity. We consistently instruct our agent to generate five unique descriptions of scenarios under each base scenario, which are then mapped into corresponding Scenic scripts for simulation. Detailed descriptions of these scenarios can be found in~\Cref{adx:scenario_description}.

\vspace{-3mm}
\paragraph{Scene.} For each route and scenario, Safebench selects approximately $9$ to $10$ of the most adversarial scenes based on testing with a SAC-trained surrogate model on each route, resulting in about $98$ to $100$ scenes for each base scenario. For our approach, the agent first generate $50$ simulations per scenario and route, updating parameter ranges every $10$ steps. From these, our agent then selects the two simulations that not only lead to a collision but also yield the lowest~\emph{\overallscore} using the same surrogate model. Consequently, this method also yields a total of $100$ scenes per base scenario, calculated as $2\times10\times5=100$. We report the average performance of all the selected scenes tested on the ego vehicles trained with three different AD algorithms for each base scenario.

\begin{table*}[t]
\small
\centering
    \caption{\small \textbf{Evaluating ego vehicle Performance Post-Finetuning:} We assess the effectiveness of various scenario generation methods based on two key metrics: \textit{collision rate} (\collisionrateabbr) and \textit{overall score} (\overallscoreabbr). For this evaluation, we finetuned the surrogate SAC-trained ego vehicle using the previously selected adversarial scenes for the first eight routes. The reported data represents the mean performance across the scenes from the last two routes, as provided by all methods. The `PP' is shorted for `Pre Pretraining,' which represents the corresponding performance of the surrogate ego vehicle on the scenes for the last two routes before finetuning. The last column provides an average across all scenarios. $\uparrow$/$\downarrow$: higher/lower the better.}
    \vspace{-3mm}
    \label{tab:adv_train}
{
\resizebox{0.8\textwidth}{!}{
\setlength{\tabcolsep}{3.75pt}
    \begin{tabular}{c|c|cccccccc|c}
    \toprule
        \multirow{3}{*}{\textbf{Metric}} & \multirow{3}{*}{\textbf{Algo.}} & \multicolumn{8}{c|}{\textbf{Base Traffic Scenarios}} & \multirow{3}{*}{\textbf{Avg.}} \\
        & & \scriptsize{\makecell{Straight \\ Obstacle}} & \scriptsize{\makecell{Turning \\ Obstacle}} & \scriptsize{\makecell{Lane \\ Changing}}  & \scriptsize{\makecell{Vehicle \\ Passing}} & \scriptsize{\makecell{Red-light \\ Running}} & \scriptsize{\makecell{Unprotected \\ Left-turn}} & \scriptsize{\makecell{Right-\\ turn}} & \scriptsize{\makecell{Crossing \\ Negotiation}} & \\
        \midrule
          \multirow{6}{*}{\collisionrateabbr $\downarrow$}  & PP & 0.48 & 0.39 & 0.58 & 0.59 & 0.69 & 0.67 & 0.46 & 0.60 & 0.559\\
          & \Learningtocollideabbr & 0.12 & 0.22 & 0.51 & 0.03 & 0.29 & \textbf{0.00} & 0.37 & 0.14 & 0.210\\
           & \AdvSimabbr  & 0.23 & 0.05 & 0.53 & \textbf{0.00} & 0.22 & 0.05 & 0.41 & 0.23 & 0.216\\
            & \CarlaGeneratorabbr & 0.22 & 0.20 & 0.39 & \textbf{0.00} & 0.04 & 0.22 & 0.19 & 0.14 & 0.176 \\
            & \AdvTrajabbr & 0.14 & 0.13 & 0.30 & \textbf{0.00} & 0.18 & \textbf{0.00} & 0.23 & 0.09 & 0.135\\
           & \name & \textbf{0.03} & \textbf{0.01} & \textbf{0.11} & 0.05 & \textbf{0.03} & 0.10 & \textbf{0.01} & \textbf{0.00} & \textbf{0.043} \\
        \midrule
          \multirow{6}{*}{\overallscoreabbr $\uparrow$} & PP & 0.673 & 0.684 & 0.648 & 0.607 & 0.609 & 0.620 & 0.651 & 0.560 & 0.632\\
          & \Learningtocollideabbr & 0.827 & 0.778 & 0.684 & 0.944 & 0.824 & 0.954 & 0.696 & 0.795 & 0.813 \\
            & \AdvSimabbr & 0.784 & 0.840 & 0.666 & \textbf{0.958} & 0.838 & 0.937 & 0.677 & 0.750 & 0.806  \\
            & \CarlaGeneratorabbr & 0.816 & 0.787 & 0.715 & 0.957 & \textbf{0.934} & 0.820 & 0.767 & 0.806 & 0.825  \\
            & \AdvTrajabbr & 0.849 & 0.783 & 0.803 & 0.955 & 0.850 & \textbf{0.948} & 0.809 & \textbf{0.915} & 0.864  \\
            & \name & \textbf{0.905} & \textbf{0.905} & \textbf{0.906} & 0.929 & \textbf{0.934} & 0.903 & \textbf{0.893} & 0.862 & \textbf{0.905} \\
        \bottomrule
    \end{tabular}
  }
}
\vspace{-6mm}
\end{table*}

\vspace{-3mm}
\paragraph{Evaluation Results:}
Our experimental results, detailed in Table~\ref{tab:test_generation}, provide a thorough evaluation of various scenario generation algorithms. These are assessed based on \textit{collision rate} (\collisionrateabbr), \textit{overall score} (\overallscoreabbr), and \emph{average displacement error} (ADE), with metrics derived from testing three distinct ego vehicle training paradigms across various base traffic scenarios. Notably, our agent, \name, consistently outperforms existing benchmarks across all metrics for each base scenario.

Specifically, \name significantly enhances the generation of safety-critical scenarios, evidenced by a marked $15\%$ increase in collision rates over the most competitive existing baselines. This substantial improvement in scenario complexity effectively challenges and evaluates autonomous driving systems in more adversarial environments. 

Regarding overall performance, our agent achieves a significant relative reduction in the overall score, amounting to $16\%$ more compared to the leading baseline. This reduction underscores the heightened complexity and challenge presented by our scenarios. Additionally, a detailed diagnostic report for the average performance on the overall eight base scenarios is provided in~\Cref{tab:diagnostic}. This report provides a detailed breakdown of the overall score across three distinct levels, encompassing a broader range of evaluations beyond the collision rate. Notably, our generated scenarios considerably diminish the average route completion rate, and they compel the ego vehicle to maintain higher average acceleration and yaw velocity, alongside frequent lane invasions, to avoid collisions with surrounding adversarial objects. These dynamics further demonstrate the effectiveness and safety-critical nature of our agent, establishing its potential to create scenarios that rigorously test autonomous driving systems. 

Moreover, \name's superiority in scenario diversity is also confirmed by achieving the highest score in ADE metrics. This outcome, indicative of the variability in adversarial objects' trajectories, highlights the comprehensive and diverse nature of our generation approach, which is essential for a thorough assessment of autonomous driving systems.

\vspace{-1mm}
In conclusion, our results demonstrate that \name not only elevates collision rates across all base traffic scenarios but also significantly lowers the overall performance scores of ego vehicles. The enhanced scenario diversity also reinforces the effectiveness of our approach. This comprehensive performance underlines the potential of our agent to set new benchmarks in the evaluation and testing of autonomous driving systems. Detailed performance for each ego vehicle trained with different AD algorithms is provided in~\Cref{apx:detailed_test_agent}.

\vspace{-2mm}
\subsection{Adversarial Training on Safety-Critical Scenarios}
\label{exp:adv}
\vspace{-2mm}

The aim of these experiments was to assess the effectiveness of safety-critical scenarios generated by various algorithms in enhancing the resilience of an ego vehicle. The findings substantiate our hypothesis that the nature of adversarial scenarios is crucial to the robustness of the ego vehicle.

\vspace{-4mm}
\paragraph{Experiment Setting.}
For maintaining the consistency, we conduct finetuning on the same surrogate SAC-trained ego vehicle under each base scenario independently, using scenes on the first eight routes generated by each algorithm, and test the adversarially finetuned ego vehicle with the selected scenes from the last two routes from all algorithms, which also resulted in around $100$ test cases for each base scenario in total. The surrogate model is finetuned with $500$ epochs, utilizing a learning rate of $0.0001$. We report the optimal performance based on evaluations conducted every $50$ epochs. Further details on finetuning settings and selecting the checkpoints can be found in~\Cref{apx:finetuning_detail}.

\vspace{-4mm}
\paragraph{Evaluation Results.}
\Cref{tab:adv_train} presents the performance outcomes of post-adversarial training, showcasing the efficacy of our agent in strengthening the robustness of the ego vehicle. Notably, when finetuned adversarially with scenarios generated by our method, the ego vehicle consistently surpassed the performance of agents trained with alternative approaches in most base scenarios. We observed a $51\%$ reduction in collision rates compared to the original ego vehicle without finetuning, and the overall score improved by $43\%$ relatively. More significantly, the collision rate was reduced by an additional $9\%$ compared to the SOTA, which indicates that our agent can effectively contribute to improving the safety and reliability of autonomous driving systems. By exposing the ego vehicle to more challenging and diverse scenarios, we are directly aiding in the advancement of robust autonomous vehicle algorithms.

In conclusion, the experimental results demonstrate the tangible benefits of our adversarial finetuning approach. The substantial reduction in collision rates, coupled with the marked enhancement in overall performance, underscores the potential of our agent in fortifying autonomous agents against adversarial perturbations. These findings signify a pivotal step towards establishing safer and more resilient autonomous driving systems, thereby fostering greater trust and reliability in real-world deployment scenarios.

\vspace{-2mm}
\section{Conclusion}
\vspace{-2mm}

In this work, we introduce \name, an LLM-based agent skilled at safety-critical scenario generation by automatically generating descriptions of safety-critical scenarios, decomposing these descriptions to retrieve the appropriate Scenic code, and subsequently compiling it to run simulations within the CARLA environment. Our experiments reveal that the scenarios produced by \name pose greater challenges, substantially elevating the collision rates for the ego vehicle under the same scenario compared to other methods. Moreover, these generated scenarios have proven to be more effective in fine-tuning the ego vehicles to avoid collisions in safety-critical situations, demonstrating the agent's utility in enhancing the robustness of autonomous vehicles.


\vspace{-2mm}
\section*{Acknowledgment}
\vspace{-2mm}
This work is partially supported by the National Science Foundation under grant No. 2046726, No. 2229876, DARPA GARD, the National Aeronautics and Space Administration (NASA) under grant No. 80NSSC20M0229, and Alfred P. Sloan Fellowship.

\clearpage

{
    \small
    \bibliographystyle{ieeenat_fullname}
    \bibliography{bib}
}

\clearpage

\appendix
\input{appendix}

\end{document}

%% file: preamble.tex
\usepackage[dvipsnames]{xcolor}
\usepackage{graphicx} 
\usepackage{url}            
\usepackage{booktabs}       
\usepackage{amsfonts}       
\usepackage{nicefrac}       
\usepackage{microtype}      
\usepackage{xcolor}         
\usepackage{xspace}
\usepackage{fancyvrb} 
\usepackage{caption}  
\usepackage{fancyvrb}
\usepackage{fvextra}
\usepackage{booktabs} 
\usepackage{enumitem}
\usepackage{amsmath}
\usepackage{caption}
\usepackage{subcaption}
\usepackage{multirow}
\usepackage{bbold}
\usepackage{makecell}
\usepackage{wrapfig}
\usepackage{minitoc}

\newcommand{\name}{\textsf{ChatScene}\xspace}
\newcommand{\blackhref}[2]{\href{#1}{\color{black}\tt \small #2}}

\newcommand{\collisionrate}{collision rate\xspace}
\newcommand{\runredlight}{frequency of running red lights\xspace}
\newcommand{\runstopsign}{frequency of running stop signs\xspace}
\newcommand{\outofroad}{average distance driven out of road\xspace}
\newcommand{\followroute}{route following stability\xspace}
\newcommand{\routecompletion}{average percentage of route completion\xspace}
\newcommand{\timespent}{average time spent to complete the route\xspace}
\newcommand{\acceleration}{average acceleration\xspace}
\newcommand{\yawvelocity}{average yaw velocity\xspace}
\newcommand{\laneinvasion}{frequency of lane invasion\xspace}
\newcommand{\overallscore}{overall score\xspace}

\newcommand{\collisionrateabbr}{CR\xspace}
\newcommand{\runredlightabbr}{RR\xspace}
\newcommand{\runstopsignabbr}{SS\xspace}
\newcommand{\outofroadabbr}{OR\xspace}
\newcommand{\followrouteabbr}{RF\xspace}
\newcommand{\routecompletionabbr}{Comp\xspace}
\newcommand{\timespentabbr}{TS\xspace}
\newcommand{\accelerationabbr}{ACC\xspace}
\newcommand{\yawvelocityabbr}{YV\xspace}
\newcommand{\laneinvasionabbr}{LI\xspace}
\newcommand{\overallscoreabbr}{OS\xspace}

\newcommand{\Learningtocollide}{Learning-to-collide\xspace}
\newcommand{\AdvSim}{AdvSim\xspace}
\newcommand{\CarlaGenerator}{Carla Scenario Generator\xspace}
\newcommand{\AdvTraj}{Adversarial Trajectory Optimization\xspace}
\newcommand{\Learningtocollideabbr}{LC\xspace}
\newcommand{\AdvSimabbr}{AS\xspace}
\newcommand{\CarlaGeneratorabbr}{CS\xspace}
\newcommand{\AdvTrajabbr}{AT\xspace}

\newcount\Comments  
\usepackage{color}
\definecolor{red}{rgb}{1,0,0}
\definecolor{darkgreen}{rgb}{0,0.5,0}
\definecolor{darkblue}{rgb}{0,0,0.5}
\definecolor{purple}{rgb}{1,0,1}
\newcommand{\kibitz}[2]{\ifnum\Comments=0\textcolor{#1}{#2}\fi}


%% file: appendix.tex
\section{Few-Shot Prompts}
\label{adx:prompt}

\subsection{Prompt for generating snippets}
\label{adx:prompt_snippet}

In this section, we present an example to illustrate our process for collecting snippets. This particular example focuses on generating more adversarial behaviors of surrounding vehicles within the base scenario of a \emph{``Straight Obstacle''}. The specific prompt employed in this process is detailed in~\Cref{tab:prompt_snippet}. Upon receiving responses from the GPT-4.0, we meticulously review each description-snippet pair for adversarial behavior. This review includes manual verification and correction of any errors, such as the use of unusable APIs, ensuring the accuracy and applicability of the snippets.

\subsection{Prompt for extracting descriptions}
\label{adx:prompt_description}
In this section, we present the few-shot prompt, as detailed in Table~\ref{tab:prompt_description}, designed for extracting component-specific descriptions from a comprehensive input description. While for extracting the specified descriptions for each component from the text, we will employ regular expressions.

\section{Detailed Scenario descriptions}
\label{adx:scenario_description}
In this section, we provide the detailed descriptions for the scenarios generated by our method for each base scenario, the full descriptions are shown in~\Cref{tab:description_1} and \Cref{tab:description_2}.

\section{Additional Experiment Details and Results}
We run all our experiments on one NVIDIA RTX A6000; we consistently use the online version of GPT-4 as the underlying LLM for manually monitoring the generation quality, and it can be easily adapted to use the API with version \texttt{gpt-4-1106-preview} for making it more automatic.

\subsection{Detailed Metric}
\label{apx:metric}
We adopt the metrics from Safebench~\cite{xu2022safebench}, and we provide the intuition for each metric here:

\begin{itemize}
    \item \collisionrateabbr (\collisionrate): Evaluates the rate of collisions, reflecting the autonomous system's accident avoidance capability.
    \item \runredlightabbr (\runredlight): Measures the frequency of running red lights, an essential aspect of traffic law adherence.
    \item \runstopsignabbr (\runstopsign): Assesses how often the vehicle fails to stop at stop signs, a key traffic rule compliance metric.
    \item \outofroadabbr (\outofroad): Quantifies the average distance the vehicle deviates from its intended roadway, indicating lane discipline.
    \item \followrouteabbr (\followroute): Examines the stability with which the vehicle follows its planned route.
    \item \routecompletionabbr (\routecompletion): Represents the average percentage of the planned route successfully completed by the vehicle.
    \item \timespentabbr (\timespent): Evaluates the time efficiency of the vehicle in completing its assigned routes.
    \item \accelerationabbr (\acceleration): Measures the average acceleration, providing insights into the smoothness of the vehicle's operation.
    \item \yawvelocityabbr (\yawvelocity): Assesses the average yaw velocity, indicating the vehicle's turning and handling characteristics.
    \item \laneinvasionabbr (\laneinvasion): Quantifies the frequency of lane invasions, a measure of lane-keeping accuracy.
\end{itemize}
While \overallscore (\overallscoreabbr) is an aggregated metric that combines all metrics above to provide a comprehensive performance overview. We also adopt the same weights in Safebench~\cite{xu2022safebench}.

\subsection{Detailed Performance for Differently Trained ego vehicles}
\label{apx:detailed_test_agent}
We provide a detailed assessment of adversarial performance in the scenes generated by different methods in each test ego vehicle (trained with SAC, PPO, TD3, respectively). Detailed statistics on the collision rate can be found in~\Cref{tab:detailed_cr}, while the overall score is reported in~\Cref{tab:detailed_os}. The results demonstrate that our method exhibits greater generalizability across diverse training algorithms for ego vehicles, consistently achieving the best average performance.

\subsection{Adversarial Finetuning Details}
We utilize a pretrained surrogate SAC-trained model from Safebench~\cite{xu2022safebench} for our experiment. The fine-tuning details are as follows:

\begin{itemize}
    \item We finetune the model with $500$ epochs (one epoch just represents one simulation for one scene).
    \item The policy learning rate and Q-value learning rate are both set at $0.0001$.
    \item The experience replay buffer is divided into two: one for non-collision cases with a size of $20,000$, and another for collision cases with a size of $200$.
    \item The batch size is fixed at $512$. In each batch, $80\%$ of the samples are drawn from the non-collision buffer, while the remaining $20\%$ are from the collision buffer. If collision cases are less than $20\% \times 512$, all available collision cases are selected.
    \item The entropy regularization coefficient, analogous to the inverse of the reward scale in the original SAC paper, is set at $0.01$.
    \item The discount factor is $0.99$, and the Q-ensemble critic comprises $2$ models.
\end{itemize}

During evaluation, it was observed that prolonged training epochs led the model to adopt a stopping strategy. Therefore, to measure the performance accurately, we assess all checkpoints post $100$ epochs at intervals of every $50$ epochs. The optimal performance is reported based on the checkpoint exhibiting the lowest collision rate while maintaining a reasonable route completion rate ($>0.3$), as adversarial events typically occur after $30\%$ completion of the total route.

\label{apx:finetuning_detail}

\begin{table*}[t]
\centering
\begin{Verbatim}[fontsize=\footnotesize, breaklines=true, breakanywhere=true, breaksymbol=, breakanywheresymbolpre=]
Your task is to provide descriptions of adversarial behaviors exhibited by surrounding objects (pedestrians, cars, cyclists, motorcycles) that may lead to a collision with the ego vehicle on a straight road. The behaviors should be safety-critical. Ensure that the provided code snippets adhere to the Scenic API without creating new APIs. Here are some refined examples:

Behavior Description: A pedestrian suddenly starts crossing the road without looking.
Snippet: 
behavior AdvBehavior():
    do CrossingBehavior(ego, globalParameters.ADV_SPEED, globalParameters.ADV_DISTANCE)
param ADV_SPEED = Range(0, 5)
param ADV_DISTANCE = Range(0, 20)

Behavior Description: A pedestrian steps onto the road right in front of the ego vehicle and stops.
Snippet:
behavior AdvBehavior():
    try:
        do CrossingBehavior(ego, globalParameters.ADV_SPEED, globalParameters.ADV_DISTANCE)
    interrupt when network.laneAt(self) is network.laneAt(ego):
        take SetWalkingSpeedAction(0)
param ADV_SPEED = Range(0, 5)
param ADV_DISTANCE = Range(0, 20)

Behavior Description: A car in an adjacent lane suddenly merges into the ego's lane.
Snippet:
behavior AdvBehavior():
    laneChangeCompleted = False
    try:
        do FollowLaneBehavior(target_speed=globalParameters.ADV_SPEED)
    interrupt when withinDistanceToAnyCars(self, globalParameters.ADV_DISTANCE) and not laneChangeCompleted:
        current_laneSection = network.laneSectionAt(self)
        leftLaneSec = current_laneSection._laneToLeft
        do LaneChangeBehavior(
            laneSectionToSwitch=leftLaneSec,
            target_speed=globalParameters.ADV_SPEED)
        laneChangeCompleted = True
param ADV_SPEED = Range(0, 10)
param ADV_DISTANCE = Range(0, 30)

Behavior Description: An adversarial cyclist sprints from behind a bus stop onto the road and stops in front of the ego vehicle.
Snippet:
behavior AdvBehavior():
    do CrossingBehavior(ego, globalParameters.OPT_ADV_SPEED, globalParameters.OPT_ADV_DISTANCE) until (distance from self to network.laneAt(ego)) < globalParameters.OPT_STOP_DISTANCE
    while True:
        take SetWalkingSpeedAction(0)
param OPT_ADV_SPEED = Range(0, 10)
param OPT_ADV_DISTANCE = Range(0, 15)
param OPT_STOP_DISTANCE = Range(0, 1)

Now, based on these examples, your task is to provide additional Scenic code snippets that simulate adversarial behaviors in traffic scenarios. Each snippet should be accompanied by a concise and clear behavior description, similar to the provided examples. Your code must follow the existing Scenic repository's API structure without introducing new APIs.
\end{Verbatim}
\caption{A few-shot prompt for generating adversarial behavior in ``Straight Obstacle'' base scenario.}
\label{tab:prompt_snippet}
\vspace{-2mm}
\end{table*}

\begin{table*}[t]
\centering
\begin{Verbatim}[fontsize=\footnotesize, breaklines=true, breakanywhere=true, breaksymbol=, breakanywheresymbolpre=]
Your task is to decompose full descriptions of safety-critical scenarios into sub-descriptions for the following distinct components:

Behavior: Describe the behavior of the adversarial object (you should also indicate the type of the object like pedestrians, cars, cyclists, and motorcycles).
Geometry: Specify the road condition where the scenario occurs (e.g., straight road, three-way intersection).
Spawn Position: Indicate the initial relative position of the adversarial object to the ego vehicle.

Here are refined examples:
Scenario: The ego vehicle is driving on a straight road, and the car in front brakes suddenly as the ego approaches.
Behavior: The adversarial car suddenly brakes when the ego approaches.
Geometry: A straight road.
Spawn Position: The adversarial car is directly in front of the ego vehicle.

Scenario: The ego vehicle attempts a right turn at a four-way intersection, and an adversarial pedestrian steps onto the road in front of it.
Behavior: The adversarial pedestrian deliberately steps onto the road right in front of the ego vehicle.
Geometry: Lanes for turning right on a four-way intersection.
Spawn Position: The adversarial pedestrian is on the right front of the ego.

Scenario: The ego vehicle navigates around a parked car, and an oncoming car suddenly turns into its path.
Behavior: The adversarial car suddenly turns into the ego's path without signaling.
Geometry: A two-lane road.
Spawn Position: The adversarial car is oncoming from the left lane of the ego.

Scenario: The ego vehicle is traveling along a straight road when a pedestrian, initially hidden behind a bus stop on the sidewalk to the right, unexpectedly dashes onto the road directly in front of the ego vehicle and comes to an abrupt stop.
Behavior: The adversarial pedestrian suddenly sprints from right, stopping abruptly in front of the ego vehicle.
Geometry: A straight road.
Spawn Position: The adversarial pedestrian is initially stationed behind a bus stop on the right front.

Scenario: The ego vehicle is changing to the right lane when an emergency vehicle approaches rapidly from behind.
Behavior: The adversarial car approaches rapidly from behind.
Geometry: A lane with right lanes on a straight road.
Spawn Position: The adversarial car is approaching from the rear on the right lane.

Scenario: The ego vehicle is turning right at an intersection, and a crossing car from left suddenly brakes.
Behavior: The adversarial car suddenly brakes near the ego vehicle.
Geometry: Lanes for turning right on an intersection
Spawn Position: The adversarial car is crossing the intersection from the left.
\end{Verbatim}
\caption{A few-shot prompt for extracting descriptions for different component given the full description of the safety-critical scenarios.}
\label{tab:prompt_description}
\end{table*}

\begin{table*}[!t]
\caption{The descriptions of scenarios generated in our method for the first four base scenarios.}
\label{tab:description_1}
\small
\centering
{
\resizebox{0.9\textwidth}{!}{
\setlength{\tabcolsep}{3.75pt}
\begin{tabular}{ccc}
\toprule
\textbf{Base Scenario} & \textbf{Description of Scenarios} \\
\midrule
\multirow{10}{*}{\makecell{Straight \\ Obstacle}}     
& \footnotesize 1. The ego vehicle is driving on a straight road; the adversarial pedestrian suddenly crosses \\
& \footnotesize the road from the right front and suddenly stops in front of the ego. \\
& \footnotesize 2. The ego vehicle is driving on a straight road; the adversarial pedestrian stands behind \\
& \footnotesize a bus stop on the right front, then suddenly sprints out onto the road in front of the ego vehicle and stops. \\
& \footnotesize 3. The ego vehicle is driving on a straight road; the adversarial pedestrian appears from a \\
& \footnotesize driveway on the left and suddenly stop and walk diagonally. \\
& \footnotesize 4. The ego vehicle is driving on a straight road; the adversarial pedestrian suddenly appears \\
& \footnotesize from behind a parked car on the right front and suddenly stop. \\
& \footnotesize 5. The ego vehicle is driving on a straight road; the adversarial pedestrian is hidden behind a vending machine \\
& \footnotesize  on the right front, and abruptly dashes out onto the road, and stops directly in the path of the ego vehicle. \\
\midrule
\multirow{10}{*}{\makecell{Turning \\ Obstacle}}      
& \footnotesize 1. The ego vehicle is turning left at an intersection; the adversarial motorcyclist on the right \\
& \footnotesize front pretends to cross the road but brakes abruptly at the edge of the road, causing confusion. \\
& \footnotesize 2. The ego vehicle is turning left at an intersection; the adversarial pedestrian on the opposite \\
& \footnotesize sidewalk suddenly crosses the road from the right front and stops in the middle of the intersection. \\
& \footnotesize 3. The ego vehicle is turning right at an intersection; the adversarial pedestrian on the left front \\
& \footnotesize suddenly crosses the road and stops in the middle of the intersection, blocking the ego vehicle's path. \\
& \footnotesize 4. The ego vehicle is turning left at an intersection; the adversarial cyclist on the left front \\
& \footnotesize suddenly stops in the middle of the intersection and dismounts, obstructing the ego vehicle's path. \\
& \footnotesize 5. The ego vehicle is turning right at an intersection; the adversarial motorcyclist on the opposite \\
& \footnotesize sidewalk abruptly crosses the road from the right front and comes to a halt in the center of the intersection. \\
\midrule
\multirow{14}{*}{\makecell{Lane \\ Changing}}
& \footnotesize 1. The ego vehicle is attempting to change lanes to avoid a slow-moving leading vehicle; \\
& \footnotesize the adversarial car in the target lane suddenly merges into the ego vehicle's original lane, \\
& \footnotesize blocking the ego from returning to its initial position. \\
& \footnotesize 2. The ego vehicle is changing to the right lane; the adversarial car is driving parallel \\
& \footnotesize to the ego and blocking its path. \\
& \footnotesize 3. The ego vehicle is performing a lane change to evade a slow-moving vehicle; the \\
& \footnotesize adversarial car in the target lane on the right front suddenly brakes, causing the ego \\
& \footnotesize vehicle to react quickly to avoid a collision. \\
& \footnotesize 4. The ego vehicle is preparing to change lanes to evade a slow-moving leading vehicle; \\
& \footnotesize the adversarial car in the target lane starts weaving between lanes, making it difficult \\
& \footnotesize for the ego to predict its position and safely execute the lane change. \\
& \footnotesize 5. The ego vehicle is attempting to change lanes to avoid a slow-moving leading vehicle; \\
& \footnotesize the adversarial car in the target lane suddenly slows down, matching the speed of the leading \\
& \footnotesize vehicle, and effectively blocks the ego from completing the lane change. \\
\midrule
\multirow{18}{*}{\makecell{Vehicle \\ Passing}}
& \footnotesize 1. The ego approaches a parked car that is blocking its lane and must use the opposite \\
& \footnotesize lane to bypass the vehicle, cautiously monitoring oncoming traffic, and suddenly encounters \\
& \footnotesize a jaywalking pedestrian, requiring the ego to quickly assess the situation and respond \\
& \footnotesize appropriately to avoid a collision. \\
& \footnotesize 2. The ego encounters a parked car blocking its lane and must use the opposite lane to \\
& \footnotesize bypass the vehicle, carefully assessing the situation and yielding to oncoming traffic, when \\
& \footnotesize an oncoming motorcyclist swerves into the lane unexpectedly, necessitating the ego to brake \\
& \footnotesize or maneuver to avoid a potential accident. \\
& \footnotesize 3. The ego approaches a parked car obstructing its lane and must use the opposite lane to go \\
& \footnotesize around when an oncoming car suddenly turns into the ego's path without signaling, requiring \\
& \footnotesize the ego to react quickly and take evasive action to prevent a collision. \\
& \footnotesize 4. The ego encounters a parked car blocking its lane and must use the opposite lane to bypass \\
& \footnotesize the vehicle when an oncoming car suddenly accelerates, closing the gap for the ego to safely \\
& \footnotesize return to its lane, necessitating the ego to quickly decide whether to accelerate or brake \\
& \footnotesize to avoid a collision. \\
& \footnotesize 5. The ego encounters a parked car blocking its lane and must use the opposite lane to bypass \\
& \footnotesize the vehicle when an oncoming pedestrian enters the lane without warning and suddenly stops, \\
& \footnotesize necessitating the ego to brake sharply or steer to avoid hitting the pedestrian. \\

 \bottomrule
\end{tabular}
}
}
\end{table*}

\begin{table*}[t]
\caption{The descriptions of scenarios generated in our method for the last four base scenarios.}
\label{tab:description_2}
\small
\centering
{
\resizebox{0.9\textwidth}{!}{
\setlength{\tabcolsep}{3.75pt}
\begin{tabular}{ccc}
\toprule
\textbf{Base Scenario} & \textbf{Description of Scenarios} \\
\midrule
\multirow{15}{*}{\makecell{Red-light \\ Running}}
& \footnotesize 1. The ego is driving straight through an intersection when a crossing vehicle runs the red light and unexpectedly\\
& \footnotesize accelerates, forcing the ego to quickly reassess the situation \\
& \footnotesize and perform a collision avoidance maneuver. \\
& \footnotesize 2. The ego vehicle is moving straight through the intersection; the adversarial vehicle, initially \\
& \footnotesize on the left front, runs the red light and makes an abrupt right turn, forcing the ego vehicle \\
& \footnotesize to perform a collision avoidance maneuver. \\
& \footnotesize 3. The ego vehicle is moving straight through the intersection; the adversarial vehicle, initially \\
& \footnotesize on the left front, runs the red light and makes an abrupt left turn, forcing the ego vehicle \\
& \footnotesize to perform a collision avoidance maneuver. \\
& \footnotesize 4. The ego moves straight at an intersection when a crossing vehicle runs the red light from \\
& \footnotesize the right and brakes abruptly, causing the ego to rapidly adapt its trajectory and perform \\
& \footnotesize a collision avoidance maneuver. \\
& \footnotesize 5. The ego vehicle is going straight through the intersection; the adversarial vehicle approaches \\
& \footnotesize from the left front and cuts off the ego vehicle. \\
\midrule
\multirow{13}{*}{\makecell{Unprotected \\ Left-turn}}
& \footnotesize 1. The ego starts an unprotected left turn at an intersection while yielding to an oncoming \\
& \footnotesize car when the oncoming car's throttle malfunctions, leading to an unexpected acceleration \\
& \footnotesize and forcing the ego to quickly modify its turning path to avoid a collision. \\
& \footnotesize 2. The ego attempts an unprotected left turn at an intersection while yielding to an oncoming \\
& \footnotesize car when the oncoming car suddenly brakes, necessitating the ego to rapidly reassess the \\
& \footnotesize situation and adjust its turn. \\
& \footnotesize 3. The ego vehicle is making an unprotected left turn; the adversarial vehicle approaches the \\
& \footnotesize intersection at a normal speed but then suddenly attempts to make a last-second right turn. \\
& \footnotesize 4. The ego attempts an unprotected left turn at an intersection while yielding to an oncoming \\
& \footnotesize car when the oncoming car veers erratically. \\
& \footnotesize 5. The ego commences an unprotected left turn at an intersection while yielding to an oncoming \\
& \footnotesize car when the adversarial car, coming from the right, blocks multiple lanes by driving extremely \\
& \footnotesize slowly, forcing the ego vehicle to change lanes. \\
\midrule
\multirow{9}{*}{\makecell{Right- \\ turn}}
& \footnotesize 1. The ego is performing a right turn at an intersection when the crossing car suddenly speeds \\
& \footnotesize up, entering the intersection and causing the ego to brake abruptly to avoid a collision. \\
& \footnotesize 2. The ego vehicle is turning right; the adversarial car (positioned ahead on the right) blocks the \\
& \footnotesize lane by braking suddenly. \\
& \footnotesize 3. The ego vehicle is turning right; the adversarial car (positioned ahead on the right) reverses abruptly. \\
& \footnotesize 4. The ego vehicle is turning right; the adversarial car (positioned behind on the right) suddenly \\
& \footnotesize accelerates and then decelerates. \\
& \footnotesize 5. The ego vehicle is turning right; the adversarial vehicle enters the intersection from the left \\
& \footnotesize side, swerving to the right suddenly. \\
\midrule
\multirow{10}{*}{\makecell{Crossing \\ Negotiation}}
& \footnotesize 1. The ego vehicle is approaching the intersection needs crossing negotiation; the adversarial car (on the left) \\
& \footnotesize suddenly accelerates and enters the intersection first and suddenly stops. \\
& \footnotesize 2. The ego vehicle is approaching the intersection needs crossing negotiation; the adversarial car (on the right) \\
& \footnotesize suddenly accelerates and enters the intersection first and suddenly stops. \\
& \footnotesize 3. The ego vehicle is entering the intersection needs crossing negotiation; the adversarial vehicle comes from \\
& \footnotesize the opposite direction and turns left and stops, causing a near collision with the ego vehicle. \\
& \footnotesize 4. The ego vehicle is entering the intersection needs crossing negotiation; the adversarial vehicle comes from\\
& \footnotesize the right and turns left and stops, causing a near collision with the ego vehicle. \\
& \footnotesize 5. The ego vehicle is entering the intersection needs crossing negotiation; the adversarial car, coming from \\
& \footnotesize the right, blocks multiple lanes by driving extremely slowly, forcing the ego vehicle to change lanes. \\
 \bottomrule
\end{tabular}
}
}
\end{table*}

\begin{table*}[ht]
\small
    \centering
    \caption{\small \textbf{Collision Rate (CR) Performance Across Different Models}. This table presents the detailed analysis of the \textit{\collisionrate} (\collisionrateabbr) for various test ego vehicles, each trained with distinct RL algorithms. We showcase the mean \collisionrateabbr for each model, demonstrating how they perform in the selected scenes under the same base scenario. Bold values denote the best performance for each scenario. Algorithms include: \Learningtocollideabbr: \Learningtocollide, \AdvSimabbr: \AdvSim, \CarlaGeneratorabbr: \CarlaGenerator, \AdvTrajabbr: \AdvTraj. Higher values of \collisionrateabbr ($\uparrow$) is preferable here.}
    \label{tab:detailed_cr}
    {
    \resizebox{0.9\textwidth}{!}{
    \setlength{\tabcolsep}{3.75pt}
    \begin{tabular}{c|c|cccccccc|c}
   \toprule
        \multirow{3}{*}{\textbf{Model}} & \multirow{3}{*}{\textbf{Algo.}} & \multicolumn{8}{c|}{\textbf{Base Traffic Scenarios}} & \multirow{3}{*}{\textbf{Avg.}} \\
        & & \scriptsize{\makecell{Straight \\ Obstacle}} & \scriptsize{\makecell{Turning \\ Obstacle}} & \scriptsize{\makecell{Lane \\ Changing}}  & \scriptsize{\makecell{Vehicle \\ Passing}} & \scriptsize{\makecell{Red-light \\ Running}} & \scriptsize{\makecell{Unprotected \\ Left-turn}} & \scriptsize{\makecell{Right-\\ turn}} & \scriptsize{\makecell{Crossing \\ Negotiation}} & \\
        \midrule
          \multirow{5}{*}{SAC (4D)} & \Learningtocollideabbr&0.40 & 0.11 & 0.60 & 0.80 & 0.92 & 0.86 & 0.70 & 0.75 & 0.642 \\
           & \AdvSimabbr& 0.53 & 0.40 & 0.75 & \textbf{0.90} & 0.62 & 0.85 & 0.21 & 0.53 & 0.599  \\
           & \CarlaGeneratorabbr &  0.53 & 0.68 & 0.67 & \textbf{0.90} & 0.76 & 0.90 & 0.69 & 0.68 & 0.725 \\
           & \AdvTrajabbr & 0.73 & 0.48 & 0.77 & \textbf{0.90} & \textbf{1.00} & \textbf{0.97} & 0.76 & \textbf{0.90} & 0.814   \\
           & \name & \textbf{0.94} & \textbf{0.73} & \textbf{0.92} & 0.81 & 0.70 & 0.88 & \textbf{0.84} & 0.78 & \textbf{0.825}  \\
        \midrule
          \multirow{5}{*}{PPO (4D)} & \Learningtocollideabbr& 0.09 & 0.11 & \textbf{1.00} & \textbf{1.00} & 0.22 & 0.20 & 0.28 & 0.00 & 0.363 \\
           & \AdvSimabbr& 0.39 & 0.19 & \textbf{1.00} & \textbf{1.00} & 0.67 & \textbf{0.61} & 0.62 & \textbf{0.92} & 0.675  \\
           & \CarlaGeneratorabbr & 0.22 & \textbf{0.61} & \textbf{1.00} & \textbf{1.00} & 0.47 & 0.37 & 0.52 & 0.44 & 0.579 \\
           & \AdvTrajabbr &0.10 & 0.11 & 0.98 & 0.87 & 0.13 & 0.21 & 0.03 & 0.05 & 0.310 \\
            & \name &\textbf{0.87} & \textbf{0.61} & 0.97 & 0.98 & \textbf{0.89} & 0.52 & \textbf{0.74} & 0.91 & \textbf{0.811} \\

  \midrule
          \multirow{5}{*}{TD3 (4D)} & \Learningtocollideabbr&0.42 & 0.06 & \textbf{1.00} & 0.70 & \textbf{1.00} & \textbf{1.00} & 0.79 & \textbf{1.00} & 0.746 \\
           & \AdvSimabbr& 0.60 & 0.39 & 0.83 & 0.70 & 0.41 & 0.65 & 0.03 & 0.26 & 0.484  \\
           & \CarlaGeneratorabbr &0.61 & 0.53 & \textbf{1.00} & 0.70 & 0.67 & 0.80 & 0.83 & 0.67 & 0.726\\
           & \AdvTrajabbr & 0.67 & 0.35 & 0.59 & 0.70 & \textbf{1.00}& 0.85 & \textbf{0.99} & 0.90 & 0.756  \\
           & \name & \textbf{0.87} & \textbf{0.75} & 0.96 & \textbf{0.99} & 0.77 & 0.86 & 0.75 & 0.90 & \textbf{0.856} \\
        \bottomrule
    \end{tabular}
    }
}
\end{table*}

\begin{table*}[ht]
\small
    \centering
    \caption{\small \textbf{Overall Score (OS) Performance Across Different Models}. This table presents the detailed analysis of the \textit{\overallscore} (\overallscoreabbr) for various test ego vehicles, each trained with distinct RL algorithms. We showcase the mean \overallscoreabbr for each model, demonstrating how they perform in the selected scenes under the same base scenario. Bold values denote the best performance for each scenario. Algorithms include: \Learningtocollideabbr: \Learningtocollide, \AdvSimabbr: \AdvSim, \CarlaGeneratorabbr: \CarlaGenerator, \AdvTrajabbr: \AdvTraj. Lower values of \overallscoreabbr ($\downarrow$) is preferable here.}
    \label{tab:detailed_os}
    {
    \resizebox{0.9\textwidth}{!}{
    \setlength{\tabcolsep}{3.75pt}
    \begin{tabular}{c|c|cccccccc|c}
   \toprule
        \multirow{3}{*}{\textbf{Model}} & \multirow{3}{*}{\textbf{Algo.}} & \multicolumn{8}{c|}{\textbf{Base Traffic Scenarios}} & \multirow{3}{*}{\textbf{Avg.}} \\
        & & \scriptsize{\makecell{Straight \\ Obstacle}} & \scriptsize{\makecell{Turning \\ Obstacle}} & \scriptsize{\makecell{Lane \\ Changing}}  & \scriptsize{\makecell{Vehicle \\ Passing}} & \scriptsize{\makecell{Red-light \\ Running}} & \scriptsize{\makecell{Unprotected \\ Left-turn}} & \scriptsize{\makecell{Right-\\ turn}} & \scriptsize{\makecell{Crossing \\ Negotiation}} & \\
        \midrule
          \multirow{5}{*}{SAC (4D)} & \Learningtocollideabbr & 0.716 & 0.824 & 0.617 & 0.515 & 0.491 & 0.521 & 0.497 & 0.500 & 0.585\\
           & \AdvSimabbr &0.663 & 0.673 & 0.552 & \textbf{0.466} & 0.650 & 0.538 & 0.746 & 0.617 & 0.613\\
           & \CarlaGeneratorabbr & 0.661 & 0.532 & 0.569 & \textbf{0.466} & 0.578 & 0.508 & 0.503 & 0.539 & 0.544\\
           & \AdvTrajabbr & 0.565 & 0.633 & 0.546 & \textbf{0.466} & \textbf{0.462} & \textbf{0.475} & 0.461 & \textbf{0.423} & 0.504 \\
           & \name & \textbf{0.450} & \textbf{0.505} & \textbf{0.451} & 0.489 & 0.582 & 0.492 & \textbf{0.426} & 0.461 & \textbf{0.482}  \\
        \midrule
          \multirow{5}{*}{PPO (4D)} & \Learningtocollideabbr & 0.858 & 0.823 & 0.457 & 0.445 & 0.850 & 0.858 & 0.702 & 0.887 & 0.735  \\
           & \AdvSimabbr & 0.726 & 0.780 & 0.457 & 0.444 & 0.617 & \textbf{0.646} & 0.535 & \textbf{0.408} & 0.577 \\
           & \CarlaGeneratorabbr & 0.806 & 0.570 & 0.468 & 0.444 & 0.722 & 0.771 & 0.582 & 0.656 & 0.627 \\
           & \AdvTrajabbr & 0.853 & 0.819 & \textbf{0.439} & 0.487 & 0.896 & 0.852 & 0.826 & 0.861 & 0.754 \\
            & \name & \textbf{0.497} & \textbf{0.578} & 0.444 & \textbf{0.421} & \textbf{0.503} & 0.705 & \textbf{0.504} & 0.417 & \textbf{0.509} \\
  \midrule
          \multirow{5}{*}{TD3 (4D)} & \Learningtocollideabbr& 0.708 & 0.843 & 0.442 & 0.561 & \textbf{0.461} & \textbf{0.466} & 0.447 & 0.378 & 0.538 \\
           & \AdvSimabbr& 0.631 & 0.668 & 0.511 & 0.561 & 0.757 & 0.638 & 0.834 & 0.755 & 0.669 \\
           & \CarlaGeneratorabbr & 0.627 & 0.599 & 0.430 & 0.561 & 0.622 & 0.559 & 0.430 & 0.542 & 0.546\\
           & \AdvTrajabbr & 0.587 & 0.689 & 0.629 & 0.561 & 0.462 & 0.534 & \textbf{0.348} & 0.423 & 0.529 \\
           & \name & \textbf{0.462} & \textbf{0.483} & \textbf{0.407} & \textbf{0.410} & 0.527 & 0.483 & 0.493 & \textbf{0.385} & \textbf{0.456} \\
        \bottomrule
    \end{tabular}
    }
}
\end{table*}